\documentclass{article}

    \PassOptionsToPackage{numbers, compress}{natbib}

    \usepackage[final]{neurips_2021}

\usepackage[utf8]{inputenc} 
\usepackage[T1]{fontenc}    
\usepackage{hyperref}       
\usepackage{url}            
\usepackage{booktabs}       
\usepackage{amsfonts}       
\usepackage{nicefrac}       
\usepackage{microtype}      
\usepackage[dvipsnames]{xcolor}
\usepackage{environ}
\usepackage{multicol}
\usepackage{multirow}
\usepackage{graphicx}
\usepackage{wrapfig}
\usepackage{tikz}
\usepackage{bbm}
\usepackage{colortbl,booktabs}
\usepackage{makecell}
\usepackage{diagbox} 
\usepackage{array}
\usepackage{subfigure}
\usepackage{amsmath, bm}
\usepackage{dsfont}
\usepackage{amssymb}
\usepackage{algorithm}
\usepackage{algpseudocode}
\usepackage{varwidth}
\usepackage{pifont}
\usepackage{makecell}
\usepackage{wrapfig}
\usepackage{enumitem}
\usepackage{caption}
\usepackage{listings}
\usepackage{color}
\definecolor{dkgreen}{rgb}{0,0.6,0}
\definecolor{gray}{rgb}{0.5,0.5,0.5}
\definecolor{mauve}{rgb}{0.58,0,0.82}

\PassOptionsToPackage{hyphens}{url}
\usepackage{hyperref}
\hypersetup{colorlinks=true}

\definecolor{Tianlong_color}{rgb}{0.858, 0.188, 0.478}

\NewEnviron{elaboration}{
\par
\begin{tikzpicture}
\node[rectangle,minimum width=0.99\textwidth] (m) {\begin{minipage}{0.99\textwidth}\BODY\end{minipage}};
\draw[thick] (m.south west) rectangle (m.north east);
\end{tikzpicture}
}

\usepackage{graphicx,calc}
\newcommand*\prune{\vcenter{\hbox{\includegraphics[width=1em]{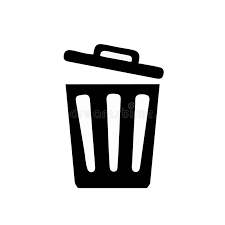}}}}

\title{Chasing Sparsity in Vision Transformers: \\ An End-to-End Exploration}

\author{%
  Tianlong Chen\textsuperscript{1}, Yu Cheng\textsuperscript{2}, Zhe Gan\textsuperscript{2}, Lu Yuan\textsuperscript{2}, \textbf{Lei Zhang\textsuperscript{3}}, \textbf{Zhangyang Wang\textsuperscript{1}}\\
  {\textsuperscript{1}University of Texas at Austin, \textsuperscript{2}Microsoft Corporation, \textsuperscript{3}International Digital Economy Academy} \\
  \small{\texttt{\{tianlong.chen,atlaswang\}@utexas.edu,\{yu.cheng,zhe.gan,luyuan\}@microsoft.com}} \\
  \small{\texttt{leizhangcn@ieee.org}} \\
}

\begin{document}

\maketitle

\begin{abstract}
Vision transformers (ViTs) have recently received explosive popularity, but their enormous model sizes and training costs remain daunting. Conventional post-training pruning often incurs higher training budgets. In contrast, this paper aims to trim down both the training memory overhead and the inference complexity, without sacrificing the achievable accuracy. We carry out the first-of-its-kind comprehensive exploration, on taking a unified approach of integrating sparsity in ViTs ``from end to end''. Specifically, instead of training full ViTs, we dynamically extract and train sparse subnetworks, while sticking to a fixed small parameter budget. Our approach jointly optimizes model parameters and explores connectivity throughout training, ending up with one sparse network as the final output. The approach is seamlessly extended from unstructured to structured sparsity, the latter by considering to guide the prune-and-grow of self-attention heads inside ViTs. We further co-explore data and architecture sparsity for additional efficiency gains by plugging in a novel learnable token selector to adaptively determine the currently most vital patches. Extensive results on ImageNet with diverse ViT backbones validate the effectiveness of our proposals which obtain significantly reduced computational cost and almost unimpaired generalization. Perhaps most surprisingly, we find that the proposed sparse (co-)training can sometimes \textit{improve the ViT accuracy} rather than compromising it, making sparsity a tantalizing ``free lunch''. For example, our sparsified DeiT-Small at ($5\%$, $50\%$) sparsity for (data, architecture), improves \textbf{$\mathbf{0.28\%}$ top-1 accuracy}, and meanwhile enjoys \textbf{$\mathbf{49.32\%}$ FLOPs} and \textbf{$\mathbf{4.40\%}$ running time} savings. Our codes are available at \url{https://github.com/VITA-Group/SViTE}. 
\vspace{-0.5em}
\end{abstract}

\section{Introduction}
\vspace{-0.5em}

Recent years have seen substantial efforts devoted to scaling deep networks to enormous sizes. Parameter counts are frequently measured in billions rather than millions, with the time and financial outlay necessary to train these models growing in concert.  The trend undoubtedly continues with the recent forefront of transformers~\cite{dosovitskiy2020image,touvron2020training,han2020survey} for computer vision tasks. By leveraging self-attention, reducing weight sharing such as convolutions, and feeding massive training data, vision transformers have established many new state-of-the-art (SOTA) records in image classification~\cite{dosovitskiy2020image,touvron2020training}, object detection~\cite{zheng2020end,carion2020end,dai2020up,zhu2021deformable}, image enhancement~\cite{chen2020pre,yang2020learning}, and image generation~\cite{parmar2018image,pmlr-v119-chen20s,jiang2021transgan}. Existing vision transformers and variants, despite the impressive empirical performance, have in general suffered from gigantic parameter-counts, heavy run-time memory usages, and tedious training. That naturally calls for the next step research of slimming their inference and training, without compromising the performance.


Model compression and efficient learning are no strangers to deep learning researchers, although their exploration in the emerging vision transformer field remains scarce \cite{zhu2021visual}. Among the large variety of compression means \cite{cheng2017survey}, sparsity has been one of the central themes since the beginning \cite{han2015deep}. 
Conventional approaches first train dense networks, and then prune a large portion of parameters in the trained networks to zero. Those methods significantly reduce the inference complexity. However, the price is to cost even more significant computational resources and memory footprints at training, since they commonly require (multiple rounds of) re-training to restore the accuracy loss~\citep{han2015deep,molchanov2019importance,zhu2017prune}. That price becomes particularly prohibitive for vision transformers, whose vanilla one-pass training is already much more tedious, slow, and unstable compared to training standard convolutional networks. 


An emerging subfield has explored the prospect of directly training smaller, sparse subnetworks in place of the full networks without sacrificing performance. The key idea is to reuse the sparsity pattern found through pruning and train a sparse network from scratch. The seminal work of lottery ticket hypothesis (LTH) \cite{frankle2018lottery} demonstrated that standard dense networks contain sparse matching subnetworks (sometimes called ``winning tickets'') capable of training in isolation to full accuracy. In other words, we could have trained smaller networks from the start if only we had known which subnetworks to choose. Unfortunately, LTH requires to empirically find these intriguing subnetworks by an iterative pruning procedure~\citep{frankle2018lottery,frankle2019linear,pmlr-v139-zhang21c,chen2020lottery,chen2020lottery2,chen2021gans,ma2021good,gan2021playing,chen2021unified,chen2021ultra}
, which still cannot get rid of the expensiveness of post-training pruning. In view of that, follow-up works reveal that sparsity patterns might emerge at the initialization \cite{lee2018snip,grasp}, the early stage of training \cite{You2020Drawing,chen2020earlybert}, or in dynamic forms throughout training \cite{mocanu2018scalable,onemillionneurons,evci2020rigging} by updating model parameters and architecture typologies simultaneously. These efforts shed light on the appealing prospect of ``end to end'' efficiency from training to inference, by involving sparsity throughout the full learning lifecycle.

This paper presents the first-of-its-kind comprehensive exploration of integrating sparsity in vision transformers (ViTs) ``from end to end''. With (dynamic) sparsity as the unified tool, we can improve the inference efficiency from both model and data perspectives, while also saving training memory costs. Our innovative efforts are unfolded along with the following three thrusts:
\begin{itemize}
    \item \textbf{From Dense to (Dynamic) Sparse:} Our primary quest is to find sparse ViTs without sacrificing the achievable accuracy, and meanwhile trimming down the training memory overhead. To meet this challenging demand, we draw inspirations from the latest sparse training  works~\citep{evci2020rigging,liu2021we} that dynamically extract and train sparse subnetworks instead of training the full models. Sticking to a fixed small parameter budget, our technique jointly optimizes model parameters and explores connectivity throughout the entire training process. We 
    term our first basic approach as \textit{Sparse Vision Transformer Exploration} (\textbf{SViTE}).
    \item \textbf{From Unstructured to Structured:} Most sparse training works~\cite{mocanu2018scalable,onemillionneurons,evci2019difficulty,mostafa2019parameter,dettmers2019sparse,liu2021selfish,dettmers2019sparse,evci2020rigging,jayakumar2020top,raihan2020sparse,liu2021we} restricted discussion to unstructured sparsity. To attain structured sparsity which is more hardware-friendly, unlike classical channel pruning available for convolutional networks, we customize a first-order importance approximation~\citep{molchanov2019importance,michel2019sixteen} to guide the prune-and-grow of self-attention heads inside ViTs. This seamlessly  extends SViTE to its second variant of \textit{Structured Sparse Vision Transformer Exploration} (\textbf{S$^2$ViTE}).
    \item \textbf{From Model to Data:} We further conduct a unified co-exploration towards joint data and architecture sparsity. That is by plugging in a novel learnable token selector to determine the most vital patch embeddings in the current input sample. The resultant framework of \textit{Sparse Vision Transformer Co-Exploration} (\textbf{SViTE+}) remains to be end-to-end trainable and can gain additional efficiency.
\end{itemize}
Extensive experiments are conducted on ImageNet with DeiT-Tiny/Small/Base. Results of substantial computation savings and nearly undamaged accuracies consistently endorse our proposals' effectiveness.
Perhaps most impressively, we find that the sparse (co-)training can even \textit{improve the ViT accuracy} rather than compromising it, making sparsity a tantalizing ``free lunch''. For example, applying SViTE+ on DeiT-Small produces superior compressed ViTs at $50\%$ model sparsity plus $5\%$ data sparsity, saving $49.32\%$ FLOPs and $4.40\%$ running time, while attaining a surprising improvement of $0.28\%$ accuracy; even when the data sparsity increases to $10\%$ (the model sparsity unchanged), there is still no accuracy degradation, meanwhile saving $52.38\%$ FLOPs and $7.63\%$ running time.

\section{Related Work}
\paragraph{Vision Transformer.} Transformer~\citep{vaswani2017attention} stems from natural language processing (NLP) applications. The Vision Transformer (ViT)~\cite{dosovitskiy2020image} pioneered to leverage a pure transformer, to encode an image by splitting it into a sequence of patches, projecting them into token embeddings, and feeding them to transformer encoders. With sufficient training data, ViT is able to outperform convolution neural networks on various image classification benchmarks~\cite{dosovitskiy2020image,guo2021cmt}. Many ViT variants have been proposed since then. For example,  DeiT~\cite{touvron2020training} and T2T-ViT~\cite{yuan2021tokens} are proposed to enhance ViT's training data efficiency, by leveraging teacher-student and better crafted architectures respectively. In addition to image classification, ViT has attracted wide attention in diverse computer vision tasks, including object detection~\cite{zheng2020end,carion2020end,dai2020up,zhu2021deformable}, segmentation~\cite{wang2020max,wang2020end},  enhancement~\cite{chen2020pre,yang2020learning}, image generation~\cite{parmar2018image,pmlr-v119-chen20s,jiang2021transgan}, video understanding~\cite{zeng2020learning,zhou2018end}, vision-language~\cite{lu2019vilbert,tan2019lxmert,chen2020uniter,su2019vl,li2019visualbert,li2020unicoder,li2020oscar,zhou2020unified} and 3D point cloud~\cite{zhao2020point}. 

Despite the impressive empirical performance, ViTs are generally heavy to train, and the trained models remain massive. That naturally motivates the study to reduce ViT inference and training costs, by considering model compression means. Model compression has been well studied in both computer vision and NLP applications~\cite{Fan2020Reducing,guo2020reweighted,ganesh2020compressing,michel2019sixteen,mccarley2019structured,chen2020lottery}. Two concurrent works~\citep{zhu2021visual,tang2021patch} made initial attempts towards ViT post-training compression by pruning the intermediate features and tokens respectively, but did not jointly consider weight pruning nor efficient training. Another loosely related field is the study of efficient attention mechanisms~\cite{zhang2021multi,parmar2018image,chen2020uniter,beltagy2020longformer,kitaev2020reformer,lee2019set,roy2021efficient,rae2019compressive,ho2019axial,katharopoulos2020transformers,choromanski2020rethinking,tay2020long,tay2020efficient,wang2020linformer}. They mainly reduce the calculation complexity for self-attention modules via various approximations such as low-rank decomposition. Our proposed techniques represent an orthogonal direction and can be potentially combined with them, which we leave as future work. Another latest concurrent work \cite{pan2021iared2} introduced an interpretable module to dynamically and gracefully drop the redundant patches, gaining not only inference efficiency but also interpretability. Being a unique and orthogonal effort from ours, their method did not consider the training efficiency yet. 



\paragraph{Pruning and Sparse Training.} Pruning is well-known to effectively reduce deep network inference costs~\citep{lecun1990optimal,han2015deep}. It can be roughly categorized into two groups: $(i)$ unstructured pruning by removing insignificant weight elements per certain criterion, such as weight magnitude~\citep{magnitude,han2015deep}, gradient~\citep{molchanov2019importance} and hessian~\citep{hessian}; $(ii)$ structured pruning~\citep{liu2017learning,he2017channel,zhou2016less} by remove model sub-structures, e.g., channels~\citep{liu2017learning,he2017channel} and attention heads~\citep{michel2019sixteen}, which are often more aligned with hardware efficiency. All above require training the full dense model first, usually for several train-prune-retrain rounds.

The recent surge of sparse training seeks to adaptively identify high-quality sparse subnetworks and train only them. Starting from scratch, those methods learn to optimize the model weights together with sparse connectivity simultaneously. \cite{mocanu2018scalable,onemillionneurons} first introduced the Sparse Evolutionary Training (SET) technique~\cite{mocanu2018scalable}, reaching superior performance compared to training with fixed sparse connectivity~\cite{mocanu2016topological,evci2019difficulty}. \cite{mostafa2019parameter,dettmers2019sparse,liu2021selfish} leverages ``weight reallocation" to improve performance of obtained sparse subnetworks. Furthermore, gradient information from the backward pass is utilized to guide the update of the dynamic sparse connectivity~\cite{dettmers2019sparse,evci2020rigging}, which produces substantial performance gains. The latest investigations ~\cite{jayakumar2020top,raihan2020sparse,liu2021we} demonstrate that more exhaustive exploration in the connectivity space plays a crucial role in the quality of found sparse subnetworks. Current sparse training methods mostly focus on convolutional networks. Most of them discuss unstructured sparsity, except a handful \cite{lym2019prunetrain,You2020Drawing} considering training convolutional networks with structured sparsity.



\section{Methodology}

\begin{figure}[t]
    \centering
    \includegraphics[width=1\linewidth]{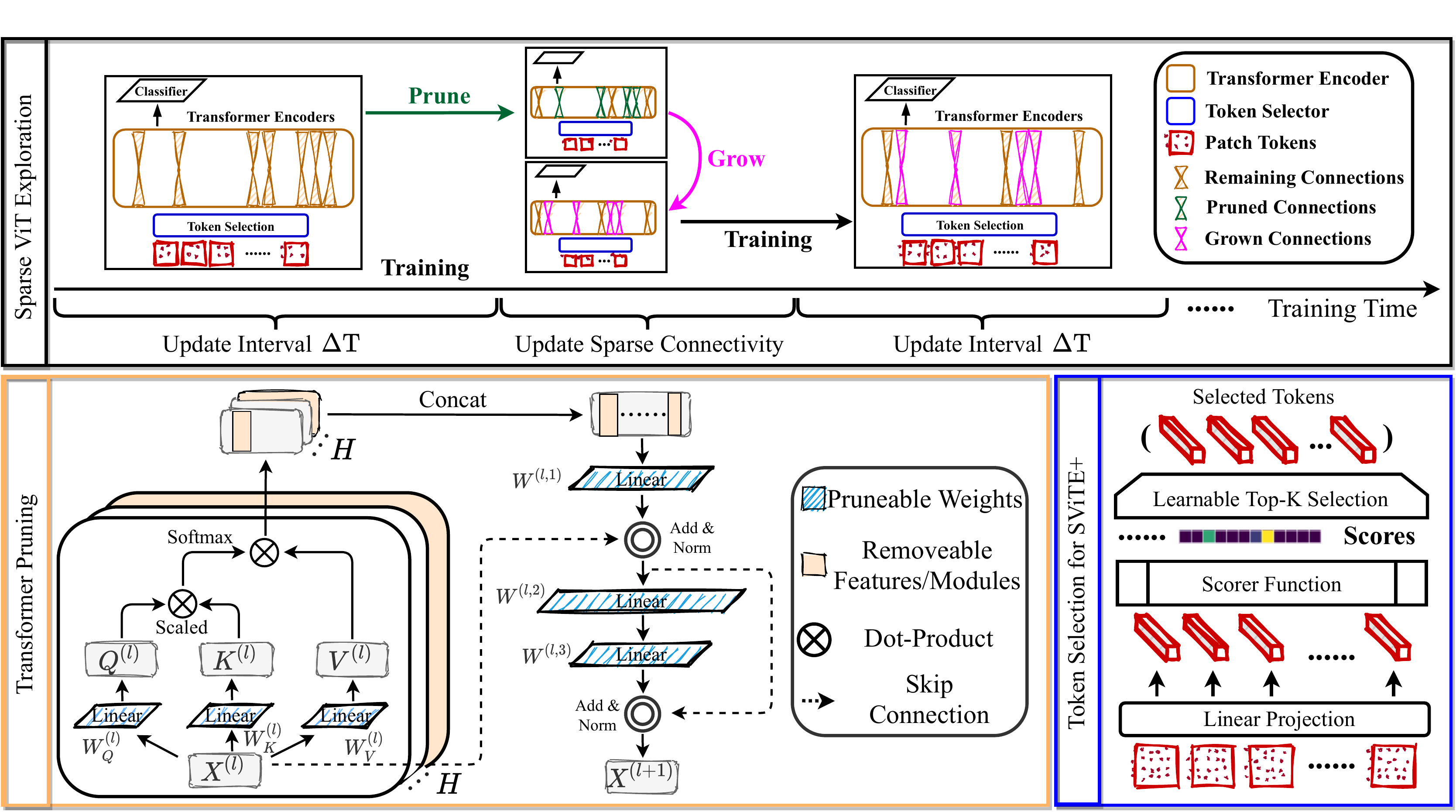}
    \vspace{-4mm}
    \caption{The overall procedure of our proposed sparse ViT exploration framework. \textit{\textbf{Upper Figure}}: first training ViT for $\Delta\mathrm{T}$ iterations, then performing prune-and-grow strategies to explore critical sparse connectivities, repreating until convergence. \textcolor{orange}{\textit{Bottom Left Figure}}: enforcing either structured or unstructured sparsity to transformer layers in ViT. \textcolor{blue}{\textit{Bottom Right Figure}}: first scoring each input embedding and applying the learnable top-$k$ selection to identify the most informative tokens.} 
    \label{fig:framework}
    \vspace{-4mm}
\end{figure}

Our SViTE method (and its variants S$^2$ViTE and SViTE+) is inspired from state-of-the-art sparse training approaches~\citep{evci2020rigging,liu2021we} in CNNs. This section presents the sparse exploration of ViT architectures, then shows the detailed procedure of input token selection for extra efficiency gains.  

\subsection{Sparse ViT Exploration} 

\paragraph{Revisiting sparse training.} Sparse training starts from a randomly sparsified model; after optimizing several iterations, it shrinks a portion of parameters based on pre-defined pruning criterion, and activates new connections w.r.t. grow indicators. After upgrading the sparse topology, it trains the new subnetwork until the next update of the connectivity. An illustration of the overall procedure is shown in Figure~\ref{fig:framework}. The key factors of sparse training are \ding{182} sparsity distribution, \ding{183} update schedule, \ding{184} pruning and \ding{185} grow criterion. 

\paragraph{Notations.} For a consistent description, we follow the standard notations in~\citep{evci2020rigging,liu2021we}. Let $\mathcal{D}$ be the training dataset. $b_t\sim\mathcal{D}$ is a randomly sampled data batch for iteration $t$. $f_W(\cdot)$ represents the model with parameters $W=(W^{(1)},\cdots,W^{(L)})$, where $W^{(l)}\in\mathbb{R}^{N_l},1\le l\le L$, $N_l$ is the number of prunable parameters in the $l_{\mathrm{th}}$ layer, and $L$ denotes the number of transformer layers. Note that the first linear projection layer and the classifier of ViT~\citep{dosovitskiy2020image,touvron2020training} are not sparsified in our framework. As illustrated in Figure~\ref{fig:framework}(bottom-left), $W_Q^{(l)}=\{W_Q^{(l,h)}\}_{h=1}^H$, $W_K^{(l)}=\{W_K^{(l,h)}\}_{h=1}^H$, $W_V^{(l)}=\{W_V^{(l,h)}\}_{h=1}^H$ are the weights of the self-attention module in the $l_{\mathrm{th}}$ layer,
$W^{(l,1)}$, $W^{(l,2)}$, $W^{(l,3)}$ are the weights of the multilayer perceptron (MLP) module in the $l_{\mathrm{th}}$ layer, and $W^{(l)}=(W_Q^{(l)},W_K^{(l)},W_V^{(l)},W^{(l,1)},W^{(l,2)},W^{(l,3)})$ collectively represent all the parameters in the $l_{\mathrm{th}}$ layer, where $H$ denotes the number of attention heads, and $1\le h\le H$. $X^{(l)}$, $Q^{(l)}$, $K^{(l)}$, and $V^{(l)}$ are the corresponding input and intermediate features, respectively. Each sparse layer only maintains a fraction $s_l\in(0,1)$ of its connections, and the overall sparsity of a sparse subnetwork is calculated as the ratio of pruned elements to the total parameter counts, i.e., $\frac{\sum_l s_l\times N_l}{\sum_l N_l}$.

\paragraph{Sparse Vision Transformer Exploration (SViTE).} SViTE explores the unstructured sparse topology in vision transformers. To be specific, we adopt \textit{Erd$\ddot{o}$s-R$\acute{e}$nyi}~\citep{mocanu2018scalable} as our \ding{182} sparsity distribution. The number of parameters in the sparse layer is scaled by $1-\frac{n_{l-1}+n_l}{n_{l-1}\times n_l}$, where $n_l$ is the number of neurons at layer $l$. This distribution allocates higher sparsities to the layers with more parameters by scaling the portion of remaining weights with the sum of the number of output and input neurons/channels. For the \ding{183} update schedule, it contains: ($i$) the update interval $\Delta\mathrm{T}$, which is the number of training iterations between two sparse topology updates; ($ii$) the end iteration $\mathrm{T}_{\mathrm{end}}$, indicating when to stop updating the sparsity connectivity, and we set $\mathrm{T}_{\mathrm{end}}$ to $80\%$ of total training iterations in our experiments; ($iii$) the initial fraction $\alpha$ of connections that can be pruned or grow, which is $50\%$ in our case; ($iv$) a decay schedule of the fraction of changeable connections $f_{\mathrm{decay}}(t,\alpha,\mathrm{T}_{\mathrm{end}})=\frac{\alpha}{2}(1+\mathrm{cos}(\frac{t\times\pi}{\mathrm{T}_{\mathrm{end}}}))$, where a cosine annealing is used, following~\citep{evci2020rigging,liu2021we}. During each connectivity update, we choose the weight magnitude as \ding{184} the pruning indicator, and gradient magnitude as \ding{185} the grow indicator. Specifically, we eliminate the parameters with the layer-wise smallest weight values by applying a binary mask $m_{\mathrm{prune}}$, then grow new connections with the highest magnitude gradients by generating a new binary mask $m_{\mathrm{grow}}$. Both masks are employed to $W^{(l)}$ via the element-wise dot product, and note that the number of non-zero elements in $m_{\mathrm{prune}}$ and $m_{\mathrm{grow}}$ are \emph{equal and fixed} across the overall procedure. Newly added connections are not activated in the last sparse topology, and are initialized to zero since it produces better performance as demonstrated in~\citep{evci2020rigging,liu2021we}.

Infrequent gradient calculation~\citep{evci2020rigging} is adopted in our case, which computes the gradients in an online manner and only stores the top gradient values. As illustrated in~\citep{evci2020rigging}, such fashion amortizes the extra effort of gradient calculation, and makes it still proportional to $1-s$ as long as $\Delta\mathrm{T}\ge\frac{1}{1-s}$, where $s$ is the overall sparsity. 

\paragraph{Structured Sparse Vision Transformer Exploration (S$^2$ViTE).} Although models with unstructured sparsity achieve superior performance, \emph{structured} sparsity~\citep{liu2017learning,he2017channel,zhou2016less} is much more hardware friendly and brings practical efficiency on realistic platforms, which motivates us to propose \emph{Structured Sparse ViT Exploration} (S$^2$ViTE). We inherit the design of \ding{182} sparsity distribution and \ding{183} update schedule from the unstructured SViTE, and a round-up function is used to eliminate decimals in the parameter counting. The key differences lie in the new \ding{184} pruning and \ding{185} grow strategies. 

\begin{wrapfigure}{r}{0.58\textwidth}\vspace{-8mm}
\begin{minipage}{0.58\textwidth}
\begin{algorithm}[H]
\caption{Sparse ViT Co-Exploration (SViTE+).}
\label{algo:DSE}
\renewcommand{\algorithmicensure}{\textbf{Initialize:}}
\begin{algorithmic}[1]
\Ensure{ViT model $f_W$, Dataset $\mathcal{D}$, Sparsity distribution $\mathbb{S}=\{s_1,\cdots,s_L\}$, Update schedule \{$\Delta\mathrm{T},\mathrm{T}_{\mathrm{end}},\alpha,f_{\mathrm{decay}}$\}, Learning rate $\eta$}
\State{Initialize $f_W$ with random sparsity $\mathbb{S}$} \Comment{\textcolor{gray}{\textit{Highly reduced parameter count.}}}
\For {each training iteration $t$}
\State \begin{varwidth}[t]{\linewidth}
Sampling a batch $b_t\sim\mathcal{D}$
\end{varwidth}
\State \begin{varwidth}[t]{0.95\linewidth}
Scoring the input token embeddings and selecting the top-$k$ informative tokens \Comment{\textcolor{gray}{\textit{Token selection}}}
\end{varwidth}
\If{($t$ mod $\Delta\mathrm{T}== 0$) and $t<\mathrm{T}_{\mathrm{end}}$} 
\For {each layer $l$}
\State{$\rho=f_{\mathrm{decay}}(t,\alpha,\mathrm{T}_{\mathrm{end}})\cdot(1-s_l)\cdot N_l$}
\State \begin{varwidth}[t]{0.8\linewidth}
Performing prune-and-grow with portion $\rho$ w.r.t. certain criterion, generating masks $m_{\mathrm{prune}}$ and $m_{\mathrm{grow}}$ to update $f_W$'s sparsity patterns \Comment{\textcolor{gray}{\textit{Connectivity exploration}}}
\end{varwidth}
\EndFor
\Else
\State{$W=W-\eta\cdot\nabla_{W}\mathcal{L}_t$} \Comment{\textcolor{gray}{\textit{Updating Weights}}}
\EndIf
\EndFor\\
\Return{a sparse ViT with a trained token selector}
\end{algorithmic}
\end{algorithm}
\end{minipage}
\vspace{-8mm}
\end{wrapfigure}

\textit{Pruning criterion}: Let $A_{(l,h)}$ denote features computed from the self-attention head \{$W_Q^{(l,h)}$, $W_K^{(l,h)}$, $W_V^{(l,h)}$\} and input embeddings $X^{(l)}$, as shown in Figure~\ref{fig:framework}. We perform the Taylor expansion to the loss function~\citep{molchanov2019importance,michel2019sixteen}, and derive a proxy score for head importance blow:
\begin{equation}
\label{eq:sorce_p}
\mathcal{I}_p^{(l,h)}=\bigg|A_{(l,h)}^{\mathrm{T}}\cdot\frac{\partial\mathcal{L}(X^{(l)})}{\partial A_{(l,h)}}\bigg|,
\end{equation}
where $\mathcal{L}(\cdot)$ is the cross-entropy loss as used in ViT. During each topology update, we remove attention heads with the smallest $\mathcal{I}_p^{(l,h)}$. For MLPs, we score neurons with $\ell_1$-norm of their associated weight vectors~\citep{bartoldson2019generalization}, and drop insignificant neurons. For example, the $j_{\mathrm{th}}$ neuron of $W^{(l,1)}$ in Figure~\ref{fig:framework} has an importance score $\|W^{(l,1)}_{j,\cdot}\|_{\ell_1}$, where $W^{(l,1)}_{j,\cdot}$ is the $j_{\mathrm{th}}$ row of $W^{(l,1)}$.

\textit{Grow criterion:} Similar to~\citep{evci2020rigging, liu2021we}, we active the new units with the highest magnitude gradients, such as $\|\frac{\partial\mathcal{L}(X^{(l)})}{\partial A_{(l,h)}}\|_{\ell_1}$ and $\|\frac{\partial\mathcal{L}(X^{(l)})}{\partial W^{(l,1)}_{j,\cdot}}\|_{\ell_1}$ for the $h_{\mathrm{th}}$ attention head and the $j_{\mathrm{th}}$ neuron of the MLP ($W^{(l,1)}$), respectively. The gradients are calculated in the same manner as the one in unstructured SViTE, and newly added units are also initialized to zero. 

\subsection{Data and Architecture Sparsity Co-Exploration for Higher Efficiency}

\begin{wrapfigure}{r}{0.63\textwidth}
    \begin{minipage}{0.63\textwidth}
\vspace{-2.2em}
        \begin{algorithm}[H]
\captionsetup{font=footnotesize}
\caption{The top-$k$ selector in a PyTorch-like style.}
\label{code:selector}\vspace{-0.5em}
\begin{lstlisting}
def topk_selector(logits, k, tau, dim=-1):
# Maintain tokens with the top-$k$ highest scores
    gumbels = -torch.empty_like(logits).exponential_().log()
    gumbels = (logits + gumbels) / tau  
    # tau is the temperature
    y_soft = gumbels.softmax(dim)
    # Straight through
    index = y_soft.topk(k, dim=dim)[1]
    y_hard = scatter(logits, index, k)
    ret = y_hard - y_soft.detach() + y_soft
    return ret
\end{lstlisting}\vspace{-0.5em}
\end{algorithm}
\vspace{-2em}
\end{minipage}
\end{wrapfigure}

Besides exploring sparse transformer architectures, we further slim the dimension of input token embeddings for extra efficiency bonus by leveraging a learnable token selector, as presented in Figure~\ref{fig:framework}. Meanwhile, the introduced data sparsity also serves as an implicit regularization for ViT training, which potentially leads to improved generalization ability, as evidenced in Table~\ref{tab:data}. Note that, due to skip connections, the number of input tokens actually determines the dimension of intermediate features, which substantially contributes to the overall computation cost. In other words, the slimmed input token embeddings directly result in compressed intermediate features, and bring substantial efficiency gains. 

For the input tokens $X^{(1)}\in\mathbb{R}^{n\times d}$, where $n$ denotes the number of tokens to be shrunk, and $d$ is the dimension of each token embedding that keeps unchanged. As shown in Figure~\ref{fig:framework}, all token embeddings are passed through a learnable scorer function which is parameterized by an MLP in our experiments. Then, a selection of the top-$k$ importance scores ($1\le k\le d$) is applied on top of it, aiming to preserve the significant tokens and remove the useless ones. To optimize parameters of the scorer function, we introduce the popular Gumbel-Softmax~\citep{gumbel,maddison2014sampling} and straight-through tricks~\cite{yin2019understanding} to enable gradient back-propagation through the top-$k$ selection, which provides an efficient solution to draw samples from a discrete probability distribution. A detailed implementation is in Algorithm~\ref{code:selector}.

The full pipeline of data and architecture co-exploration is summarized in Algorithm~\ref{algo:DSE}. We term this approach SViTE+. We first feed the randomly sampled data batch to the token selector and pick the top-$k$ informative token embeddings. Then, we alternatively train the sparse ViT for $\Delta\mathrm{T}$ iterations and perform prune-and-grow to explore the sparse connectivity in ViTs dynamically. In the end, a sparse ViT model with a trained token selector is returned and ready for evaluation.
\vspace{-2mm}
\section{Experiments} \label{sec:exp}
\vspace{-2mm}


\paragraph{Baseline pruning methods.} We extend several effective pruning methods from CNN compression as our strong baselines. \textit{Unstructured pruning}: $(i)$ One-shot weight Magnitude Pruning (OMP)~\citep{han2015deep}, which removes insignificant parameters with the globally smallest weight values; $(ii)$ Gradually Magnitude Pruning (GMP)~\citep{zhu2017prune}, which seamlessly incorporates gradual pruning techniques within the training process by eliminating a few small magnitude weights per iteration; and $(iii)$ Taylor Pruning (TP)~\cite{molchanov2019importance}, which utilizes the first-order approximation of the training loss to estimate units' importance for model sparsification. \textit{Structured pruning:} Salience-based Structured Pruning (SSP). We draw inspiration from~\citep{michel2019sixteen,bartoldson2019generalization}, and remove sub-modules in ViT (e.g., self-attention heads) by leveraging their weight, activation, and gradient information. Moreover, due to the repetitive architecture of ViT, we can easily reduce the number of transformer layers to create a smaller dense ViT (Small-Dense) baseline that has similar parameter counts to the pruned ViT model. 

\begin{wrapfigure}{r}{0.52\linewidth}
\centering
\vspace{-7mm}
\includegraphics[width=1\linewidth]{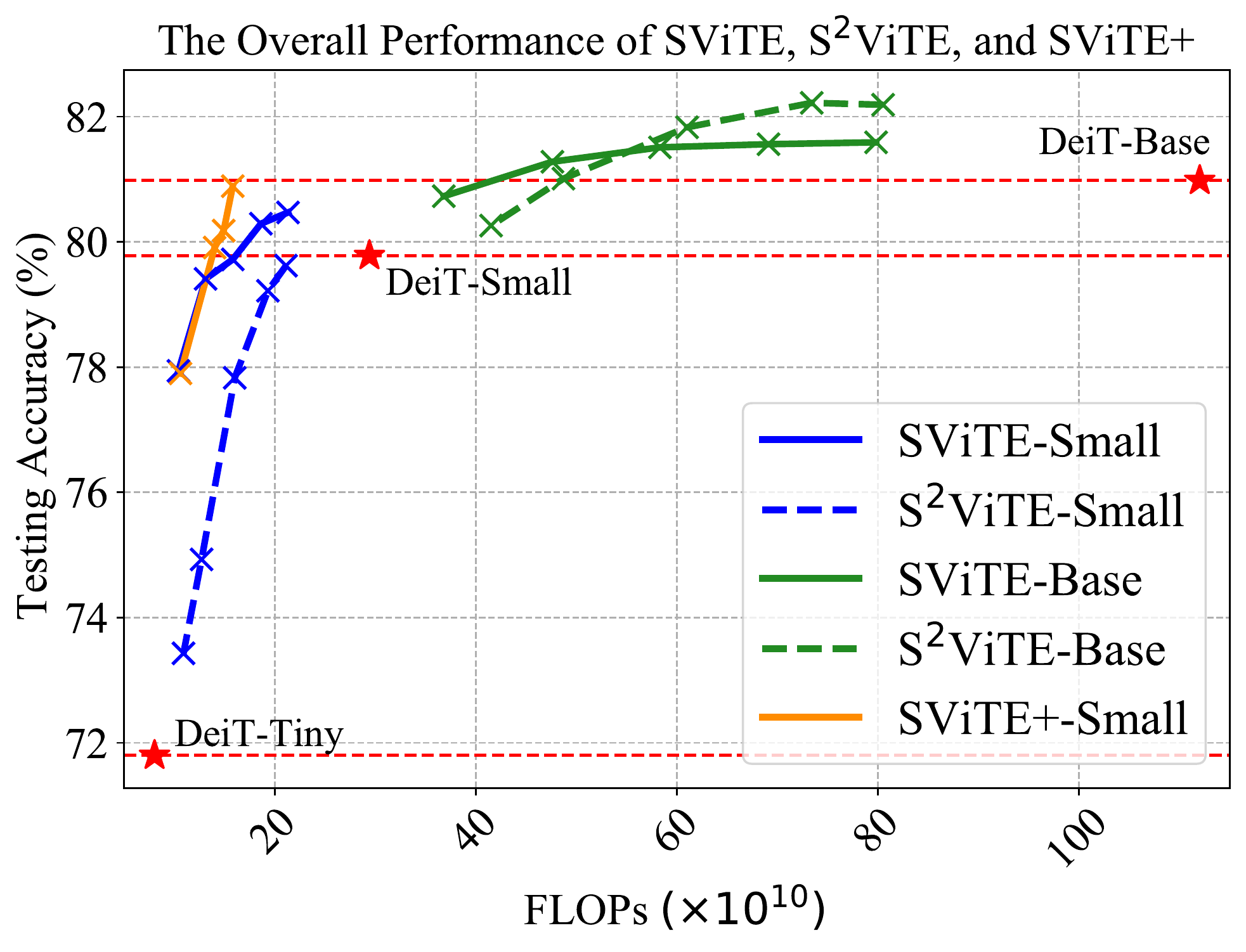}
\vspace{-7mm}
\caption{\small Top-1 accuracy (\%) over FLOPs ($\times10^{10}$) on ImageNet of our methods, i.e., SViTE, S$^2$ViTE, and SViTE+, compared to DeiT baselines, trained on Imagenet-1K only.}
\label{fig:dse_overall}
\vspace{-8mm}
\end{wrapfigure}

\vspace{-2mm}
\paragraph{Implementation details.} Our experiments are conducted on ImageNet with DeiT-Tiny/Small/Base backbones. The detailed training configurations are listed in Table~\ref{tab:setup}, which mainly follows the default setups in~\citep{touvron2020training}. All involved customized hyperparameters are tuned via grid search (later shown in Figure~\ref{fig:ablation}). For a better exploration of sparsity connectivities, we increase training epochs to 600 for all experiments. GMP~\citep{zhu2017prune} has an additional hyperparameter, i.e., the pruning schedule, which starts from $\frac{1}{6}$ and ends at $\frac{1}{2}$ of the training epochs with $20$ times pruning in total. More details are referred to Appendix~\ref{sec:more_implementation}.

\vspace{-2mm}
\paragraph{Training time measuring protocol.} We strictly measure the running time saving of (sparse) vision transformers on the ImageNet-1K task using CUDA benchmark mode. To be specific, we separately calculate the time elapsed during each iteration, to eliminate the impact of the hardware environment as much as possible. Note that the time for the data I/O is excluded.

\begin{table}[t]
\caption{Details of training configurations in our experiments, mainly following the settings in~\citep{touvron2020training}.}
\label{tab:setup}
\centering
\resizebox{1\textwidth}{!}{
\begin{tabular}{l|ccc|c}
\toprule
\multirow{1}{*}{Backbone} & \multirow{1}{*}{Update Schedule $\{\Delta\mathrm{T},\mathrm{T}_{\mathrm{end}},\alpha,f_{\mathrm{decay}}\}$} & \multirow{1}{*}{Batch Size} & \multirow{1}{*}{Epochs} & \multirow{1}{*}{Inherited Settings from DeiT~\citep{touvron2020training}}\\  
\midrule
DeiT-Tiny & $\{20000,1200000,0.5,\mathrm{cosine}\}$ & 512 & 600 & \multirow{3}{*}{\begin{tabular}[c]{@{}c@{}} AdamW, $0.0005\times\frac{\mathrm{batchsize}}{512}$, cosine decay \\ warmup 5 epochs, 0.05 weight decay \\ 0.1 label smoothing, augmentations, etc. \end{tabular}}\\ 
DeiT-Small & $\{15000,1200000,0.5,\mathrm{cosine}\}$ & 512 & 600 & \\
DeiT-Base & $\{7000,600000,0.5,\mathrm{cosine}\}$ & 1024 & 600 & \\
\bottomrule
\end{tabular}}
\vspace{-5mm}
\end{table}

\vspace{-2mm}
\paragraph{Highlight of our findings.} The overall performance of SViTE, S$^2$ViTE, and SViTE+ on DeiT backbones are summarized in Figure~\ref{fig:dse_overall}. We highlight some  takeaways below. \par

\begin{elaboration}
\textbf{Takeaways:} \ding{182} SViTE produces sparse DeiTs with enhanced generalization and substantial reduced FLOPs, compared to its dense counterpart (\textcolor{red}{$\star$}). SViTE+ further improves the performance of SViTE by selecting the most vital patches. \ding{183} S$^2$ViTE achieves matched accuracy on DeiT-Small, and significantly enhances performance on DeiT-Base. Meanwhile, its structural sparsity brings considerable running time savings. \ding{184} Appropriate data and architecture sparsities can effectively regularize ViT training, leading to a new SOTA win-win between ViT accuracy and efficiency. 
\end{elaboration}

\subsection{SViTE with Unstructured Sparsity}

We perform SViTE to mine vital unstructured sparsity in DeiTs~\citep{touvron2020training}. Solid lines in Figure~\ref{fig:dse_overall} record the top-1 test-set accuracy over FLOPs on ImageNet-1K of SViTE-Small and SViTE-Base with a range of sparsity from $30\%$ to $70\%$. In general, we observe that SViTE generates superior sparse ViTs with both accuracy and efficiency gains. Table~\ref{tab:unstructure_tiny},~\ref{tab:unstructure_small}, and~\ref{tab:unstructure_base} present the comparison between SViTE and various pruning baselines. From these extensive results, we draw several consistent observations. \underline{First}, compared to the dense baselines, SViTE-Tiny, -Small, and -Base obtain $25.56\%\sim34.16\%$, $46.26\%\sim55.44\%$, and $47.95\%\sim57.50\%$ FLOPs reduction, respectively, at $30\%\sim60\%$ sparsity levels with only a negligible accuracy drop within $0.5\%$. It verifies the effectiveness of our proposal, and indicates severe parameter redundancy in ViT. \underline{Second}, our SViTE models from dynamic explorations consistently surpass other competitive baseline methods, including OMP, GMP, TP, and Small-Dense by a substantial performance margin. Among all the baseline approaches, GMP that advocates a gradual pruning schedule achieves the best accuracy with all three DeiT backbones. \underline{Third}, in Figure~\ref{fig:dse_overall}, both SViTE-Small (\textcolor{blue}{blue} solid line) and SViTE-Base (\textcolor{green}{green} solid line) show an improved trade-off between accuracy and efficiency, compared to their dense DeiT counterparts. Interestingly, we also observe that with similar parameter counts, a large sparse ViT consistently outperforms the corresponding smaller dense ViT. A possible explanation is those appropriate sparse typologies regularize network training and lead to enhanced generalization, which coincides with recent findings of critical subnetworks (i.e., winning tickets) in dense CNNs~\citep{frankle2019lottery,chen2020lottery2} and NLP transformer~\citep{chen2020lottery,gale2019state} models.

\vspace{-4mm}
\begin{multicols}{2}
\begin{table}[H]
\caption{\small Results of \textbf{SViTE-Tiny} on ImageNet-1K. Accuracies (\%) within/out of parenthesis are the reproduced/reported~\citep{touvron2020training} performance.}
\label{tab:unstructure_tiny}
\centering
\resizebox{0.49\textwidth}{!}{
\begin{tabular}{l|cc|c}
\toprule
\multirow{1}{*}{Models} & \multirow{1}{*}{Sparsity (\#Para.)} & \multirow{1}{*}{FLOPs Saving} & \multirow{1}{*}{Accuracy (\%)}  \\  
\midrule
DeiT-Tiny & 0\% (5.72M) & 0\% & 72.20 \scalebox{0.8}{(71.80)} \\ \midrule
\rowcolor[gray]{0.9} 
SViTE-Tiny & 30\% (4.02M)  & 25.56\% & 71.78\\
OMP & 30\% (4.02M) & 25.56\% & 68.35 \\
GMP & 30\% (4.02M) &  25.56\% & 69.56 \\
TP & 30\% (4.02M) &  25.56\% & 68.38 \\ \midrule
\rowcolor[gray]{0.9} 
SViTE-Tiny & 40\% (3.46M) & 34.16\% & 71.75 \\
OMP & 40\% (3.46M) &  34.16\% & 66.52\\
GMP & 40\% (3.46M) & 34.15\% & 68.36 \\
TP & 40\% (3.46M) & 34.17\% &  65.45\\ \midrule
Small-Dense & 0\% (3.94M) & 32.54\% & 67.33 \\ 
\bottomrule
\end{tabular}}
\end{table}

\begin{table}[H]
\caption{\small Results of \textbf{SViTE-Small} on ImageNet-1K. Accuracies (\%) within/out of parenthesis are the reproduced/reported~\citep{touvron2020training} performance.}
\label{tab:unstructure_small}
\centering
\resizebox{0.49\textwidth}{!}{
\begin{tabular}{l|cc|c}
\toprule
\multirow{1}{*}{Models} & \multirow{1}{*}{Sparsity (\#Para.)} & \multirow{1}{*}{FLOPs Saving} & \multirow{1}{*}{Accuracy (\%)}  \\  
\midrule
DeiT-Small & 0\% (22.1M) & 0\% & 79.90 \scalebox{0.8}{(79.78)} \\ \midrule
\rowcolor[gray]{0.9} 
SViTE-Small & 50\% (11.1M) & 46.26\% & 79.72 \\
OMP & 50\% (11.1M) & 46.25\% & 76.32 \\
GMP & 50\% (11.1M) & 46.26\% & 76.88 \\
TP & 50\% (11.1M) & 46.26\% & 76.30 \\ \midrule
\rowcolor[gray]{0.9} 
SViTE-Small & 60\% (8.9M) & 55.44\% & 79.41 \\
OMP & 60\% (8.9M) & 55.44\% & 75.32 \\
GMP & 60\% (8.9M) & 55.44\% & 76.79 \\
TP & 60\% (8.9M) & 55.44\% & 74.50 \\ \midrule
Small-Dense & 0\% (11.4M) & 49.32\% & 73.93 \\ 
\bottomrule
\end{tabular}}
\end{table}

\end{multicols}

\vspace{-8mm}
\begin{table}[H]
\caption{\small Results of S$^2$ViTE with structured sparsity on ImageNet-1K with DeiT-Tiny/Small/Base. Accuracies (\%) within/out of parenthesis are the reproduced/reported~\citep{touvron2020training} performance.}
\label{tab:structure}
\centering
\resizebox{1\textwidth}{!}{
\begin{tabular}{l|cccc|c}
\toprule
\multirow{1}{*}{Models} & \multirow{1}{*}{Sparsity (\%)} & \multirow{1}{*}{Parameters} & \multirow{1}{*}{FLOPs Saving} & \multirow{1}{*}{Running Time Reduced} & \multirow{1}{*}{Top-1 Accuracy (\%)}  \\  
\midrule
DeiT-Tiny (Dense) & 0\% & 5.72M & 0\% & 0\%  & 72.20 \scalebox{0.8}{(71.80)}\\ 
SViTE-Tiny (Unstructured) & 30\% & 4.02M & 25.56\% & 0\% & 71.78\\ 
SSP-Tiny (Structured) & 30\% & 4.21M & 23.69\% & 10.57\% & 68.59\\
\rowcolor[gray]{0.9} 
S$^2$ViTE-Tiny (Structured) & 30\% & 4.21M & 23.69\% & 10.57\% & 70.12 \\ \midrule
DeiT-Small (Dense) & 0\% & 22.1M & 0\% & 0\% & 79.90 \scalebox{0.8}{(79.78)}\\ 
SViTE-Small (Unstructured) & 40\% & 13.3M & 36.73\% & 0\% & 80.26 \\ 
SSP-Small (Structured) & 40\% & 14.6M & 31.63\% & 22.65\% & 77.74\\
\rowcolor[gray]{0.9} 
S$^2$ViTE-Small (Structured) & 40\% & 14.6M & 31.63\% & 22.65\% & 79.22\\ \midrule
DeiT-Base (Dense) & 0\% & 86.6M & 0\% & 0\% &  81.80 \scalebox{0.8}{(80.98)}\\ 
SViTE-Base (Unstructured) & 40\% & 52.0M & 38.30\% & 0\% & 81.56\\
SSP-Base (Structured) & 40\% & 56.8M & 33.13\% & 24.70\% & 80.08\\
\rowcolor[gray]{0.9} 
S$^2$ViTE-Base (Structured) & 40\% & 56.8M & 33.13\% & 24.70\% & 82.22 \\ 
\bottomrule
\end{tabular}}
\vspace{-4mm}
\end{table}

\subsection{S$^2$ViTE with Structured Sparsity}
For more practical benefits, we investigate sparse DeiTs with structured sparsity. Results are summarized in Table~\ref{tab:structure}. Besides the obtained $23.79\%\sim33.63\%$ FLOPs savings, S$^2$ViTE-Tiny, S$^2$ViTE-Small, and S$^2$ViTE-Base enjoy an extra $10.57\%$, $22.65\%$, and $24.70\%$ running time reduction, respectively, from $30\%\sim40\%$ structured sparsity with competitive top-1 accuracies. Furthermore, S$^2$ViTE consistently outperforms the baseline structured pruning method (SSP), which again demonstrates the superior sparse connectivity learned from dynamic sparse training. 

The most impressive results come from S$^2$ViTE-Base at $40\%$ structured sparsity. It even surpasses the dense DeiT base model by $0.42\%\sim1.24\%$ accuracy with $34.41\%$ parameter counts, $33.13\%$ FLOPs, and $24.70\%$ running time reductions. We conclude that ($i$) an adequate sparsity from S$^2$ViTE boosts ViT's generalization ability, which can be regarded as an implicit regularization; ($ii$) larger ViTs (e.g., DeiT-Base) tend to have more superfluous self-attention heads, and are more amenable to structural sparsification from S$^2$ViTE, based on Figure~\ref{fig:dse_overall} where dash lines denote the overall performance of S$^2$ViTE-Small and S$^2$ViTE-Base with a range of sparsity from $30\%$ to $70\%$.

\vspace{-4mm}
\begin{multicols}{2}

\begin{table}[H]
\caption{\small Results of \textbf{SViTE-Base} on ImageNet-1K. Accuracies (\%) within/out of parenthesis are the reproduced/reported~\citep{touvron2020training} performance.}
\label{tab:unstructure_base}
\vspace{0.5mm}
\centering
\resizebox{0.50\textwidth}{!}{
\begin{tabular}{l|cc|c}
\toprule
\multirow{1}{*}{Models} & \multirow{1}{*}{Sparsity (\#Para.)} & \multirow{1}{*}{FLOPs Saving} & \multirow{1}{*}{Accuracy (\%)}  \\   
\midrule
DeiT-Base & 0\% (86.6M) & 0\% & 81.80 \scalebox{0.8}{(80.98)}\\ \midrule
\rowcolor[gray]{0.9} 
SViTE-Base & 50\% (43.4M) & 47.95\% & 81.51 \\
OMP & 50\% (43.4M) & 47.94\% & 80.26 \\
GMP & 50\% (43.4M) & 47.95\% & 80.79 \\
TP & 50\% (43.4M) & 47.94\% & 80.55 \\ \midrule
\rowcolor[gray]{0.9} 
SViTE-Base & 60\% (34.8M) & 57.50\% & 81.28 \\
OMP & 60\% (34.8M) & 57.50\% & 80.25 \\
GMP & 60\% (34.8M) & 57.50\% & 80.44 \\
TP & 60\% (34.8M) & 57.49\% & 80.37 \\ \midrule
Small-Dense & 0\% (44.0M) &  49.46\% & 78.59\\ 
\bottomrule
\end{tabular}}
\end{table}

\begin{table}[H]
\caption{\small Results of \textbf{SViTE+-Small} on ImageNet-1K. Accuracies (\%) within/out of parenthesis are the reproduced/reported~\citep{touvron2020training} performance.}
\label{tab:data}
\centering
\resizebox{0.49\textwidth}{!}{
\begin{tabular}{c|cc|c}
\toprule
\multirow{1}{*}{\#Tokens (\%)} & \multirow{1}{*}{Time Reduced} & \multirow{1}{*}{FLOPs Saving} & \multirow{1}{*}{Accuracy (\%)}  \\  
\midrule
\multicolumn{4}{c}{SViTE+-Small 50\% Unstructured Sparsity}\\ \midrule
100\% & 0\% & 46.26\% & 79.72 \\
\rowcolor[gray]{0.9} 
95\% & 4.40\% & 49.32\% & 80.18 \\
\rowcolor[gray]{0.9} 
90\% & 7.63\% & 52.38\% & 79.91 \\
\rowcolor[gray]{0.9} 
70\% & 19.77\% & 63.95\% & 77.90 \\ \midrule
\multicolumn{4}{c}{S$^2$ViTE+-Small 40\% Structured Sparsity}\\ \midrule
100\% & 22.65\% & 31.63\% & 79.22 \\
\rowcolor[gray]{0.9} 
95\% & 27.17\% & 37.76\% & 78.44 \\
\rowcolor[gray]{0.9} 
90\% & 29.21\% & 41.50\% & 78.16 \\
\rowcolor[gray]{0.9} 
70\% & 39.10\% & 54.96\% & 74.77 \\ 
\bottomrule
\end{tabular}}
\end{table}

\end{multicols}
\vspace{-7mm}

\subsection{SViTE+ with Data and Architecture Sparsity Co-Exploration}  \label{sec:data_model}
In this section, we study data and architecture sparsity co-exploration for ViTs, i.e., SViTE+. Blessed by the reduced input token embeddings, even ViTs with unstructured sparsity can have running time savings. The benefits are mainly from the shrunk input and intermediate feature dimensions. Without loss of generality, we consider SViTE+-Small with $50\%$ unstructured sparsity and S$^2$ViTE+-Small with $40\%$ structured sparsity as examples. As shown in Table~\ref{tab:data} and Figure~\ref{fig:dse_overall}, SViTE+-Small at $50\%$ unstructured sparsity is capable of abandoning $5\%\sim10\%$ tokens while achieving $4.40\%\sim7.63\%$ running time and $49.32\%\sim52.38\%$ FLOPs savings, with even improved top-1 testing accuracy. It again demonstrates that data sparsity as an implicit regularizer plays a beneficial role in ViT training. However, slimming input and intermediate embedding is less effective when incorporated with S$^2$ViTE, suggesting that aggressively removing structural sub-modules hurts ViT's generalization.

\subsection{Ablation and Generalization Study of SViTEs}

\begin{figure}[!ht]
    \centering
    \vspace{-2mm}
    \includegraphics[width=0.98\linewidth]{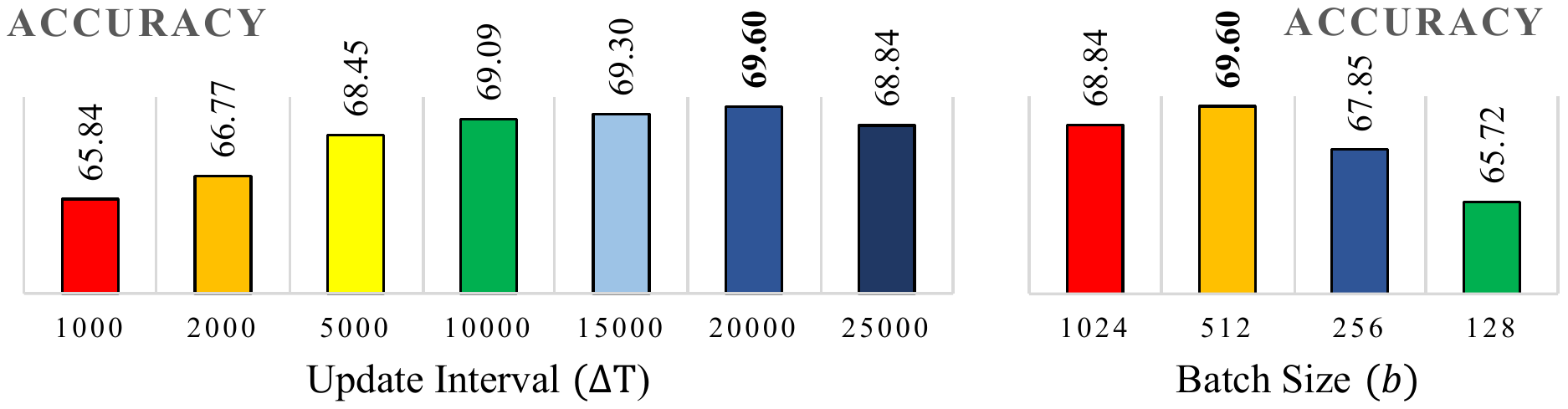}
    \vspace{-2mm}
    \caption{\small Accuracy of SViTE-Tiny with $50\%$ unstructured sparsity. \textit{Left}: ablation studies of the update interval $(\Delta\mathrm{T})$; \textit{Right}: ablations studies of the adopted batch size $(b)$.}
    \vspace{-4mm}
    \label{fig:ablation}
\end{figure}

\paragraph{Update interval in SViTE.} The length of the update interval $\Delta\mathrm{T}$ controls one of the essential trade-offs in our proposed dynamic sparse exploration, since $\Delta\mathrm{T}$ multiplying the number of updates is the pre-defined $\mathrm{T}_{\mathrm{end}}$. On the one hand, a larger updated interval (i.e., smaller update frequency) produces a more well-trained model for improved estimation of units' importance. On the other hand, a larger update frequency (i.e., smaller $\Delta\mathrm{T}$) allows more sufficient exploration of sparse connectivities, which potentially generates higher-quality sparse subnetworks, as demonstrated in~\citep{liu2021we}. We evaluate this factor in our SViTE context, and collect the results in Figure~\ref{fig:ablation} (\textit{Left}). We observe that $\Delta\mathrm{T}=20000$ works the best for SViTE-Tiny, and both larger and smaller $\Delta\mathrm{T}$ degrade the performance.

\paragraph{Batch size in SViTE.} Besides the update interval $\Delta\mathrm{T}$, batch size ($b$) also affects the aforementioned trade-off, especially for the data-hungry ViT training. We investigate different batch sizes in Figure~\ref{fig:ablation} (\textit{Right}), and find that $b=512$ outperforms other common options for SViTE-Tiny.

\paragraph{Generalization study of SViTE and its variants.} It is worth mentioning that our proposed frameworks (SViTE, S$^2$ViTE, SViTE+) are independent of the backbone architectures, and can be easily plugged in other vision transformer models~\cite{han2021transformer,yuan2021tokens,wang2021pyramid,zhou2021deepvit}. We implemented both SViTE and S$^2$ViTE on TNT-S~\cite{han2021transformer}. SViTE-TNT-S gains $0.13$ accuracy improvements (Ours: $81.63$ v.s. TNT-S: $81.50$) and $37.54\%$ FLOPs savings at $40\%$ unstructured sparsity; S$^2$ViTE-TNT-S obtains $32.96\%$ FLOPs and $23.71\%$ running time reductions at $40\%$ structured sparsity with almost unimpaired accuracy (Ours: $81.34$ v.s. TNT-S:$81.50$). 

\begin{figure}[t]
    \centering
    \includegraphics[width=1\linewidth]{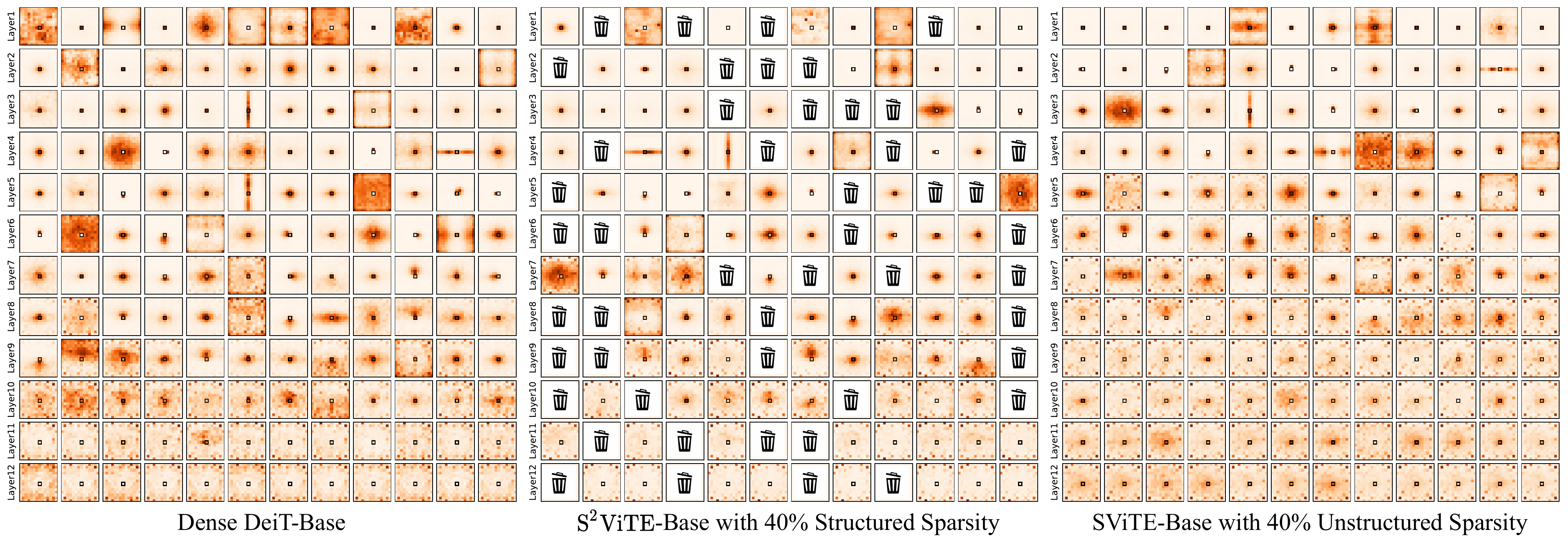}
    \vspace{-6mm}
    \caption{\small Attention probabilities for DeiT-Base, S$^2$ViTE-Base, and SViTE-Base models with $12$ layers (rows) and $12$ heads (columns) using visualization tools provided in~\citep{Cordonnier2020On}. Attention maps are averaged over $100$ test samples from ImageNet-1K to present head behavior and remove the dependence on the input content. The black square is the query pixel. $\prune$ indicates pruned attention heads. Zoom-in for better visibility.}
    \label{fig:attention}
    \vspace{-4mm}
\end{figure}

\begin{figure}[t]
    \centering
    \includegraphics[width=1\linewidth]{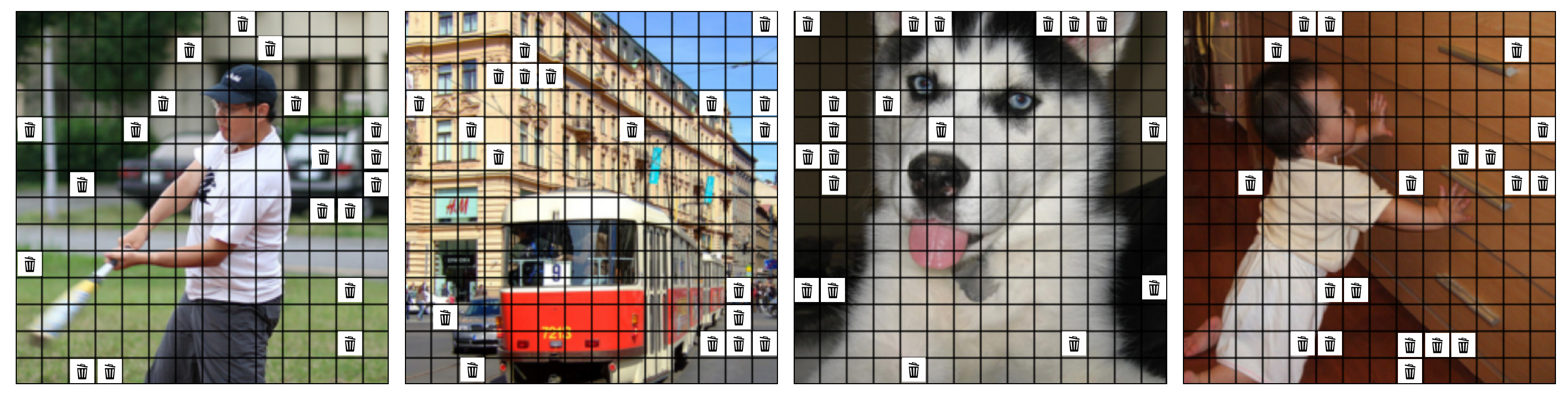}
    \vspace{-7mm}
    \caption{\small Learned patch selection patterns of SViTE+-Small at $10\%$ data and $50\%$ architecture sparsity levels. $\prune$ indicates removed inessential patches.}
    \label{fig:patch}
    \vspace{-4mm}
\end{figure}

\subsection{Visualization}
\paragraph{Sparse connectivity patterns.} We provide unit-wise and element-wise heatmap visualizations for SViTE-Base with $40\%$ structured sparsity in Figure~\ref{fig:mask_s} (in Appendix). Similarly, element-wise heatmap visualizations of SViTE-Base with $50\%$ unstructured sparsity are displayed in Figure~\ref{fig:mask_uns}. We find that even unstructured sparsity exploration can develop obvious structural patterns (i.e., ``vertical lines'' in mask heatmaps), which implies a stronger potential for hardware speedup~\citep{elsen2020fast}.

\vspace{-1mm}
\paragraph{Self-attention heatmaps.} As shown in Figure~\ref{fig:attention}, we utilize tools in~\citep{Cordonnier2020On} to visualize attention maps of (sparse) ViTs. Multiple attention heads show similar behaviors, which implies the structural redundancy. Fortunately, S$^2$ViTE eliminates unnecessary heads to some extent. With regard to SViTE-Base's visual results, it seems to activate fewer attention heads for predictions (darker colors mean larger values), compared to the ones of dense DeiT-Base. We also observe that in the bottom layers, the attention probabilities are more centered at several heads; while in the top layers, the attention probabilities are more uniformly distributed. This kind of tendency is well preserved by our sparse ViT (SViTE) from Dense ViTs.

\vspace{-1mm}
\paragraph{Learned patch selection patterns.} Figure~\ref{fig:patch} presents the learned behaviors of our token selector in SViTE+. We observe that the useless removed patches are typically distributed around the main object or in the background. Meanwhile, the patches within the objects of interest are largely persevered, which evidences the effectiveness of our learned patch token selector.

\vspace{-1mm}
\section{Conclusion and Discussion of Broader Impact} \label{sec:conclusion}
\vspace{-1mm}
In this work, we introduce sparse ViT exploration algorithms, SViTE, and its variants S$^2$ViTE and SViTE+, to explore high-quality sparse patterns in both ViT's architecture and input token embeddings, alleviating training memory bottleneck and pursuing inference ultra-efficiency (e.g., running time and FLOPs). Comprehensive experiments on ImageNet validate the effectiveness of our proposal. Our informative visualizations further demonstrate that SViTE+ is capable of mining crucial connections and input tokens by eliminating redundant units and dropping useless token embeddings. Future work includes examining the performance of our sparse ViTs on incoming hardware accelerators~\citep{8465793,ashbyexploiting,liu2018memory,7551397,8686088}, which will provide better supports for sparsity. 

This work is scientific in nature, and we do not believe it has immediate negative societal impacts. Our findings of sparse vision transformers are highly likely to reduce both memory and energy costs substantially, leading to economic deployment in real-world applications (e.g., on smartphones).

\vspace{-0.5em}
\section*{Acknowledgment}
\vspace{-0.5em}
Z.W. is in part supported by an NSF RTML project (\#2053279).


\bibliographystyle{unsrt}
\bibliography{ref}


\clearpage

\appendix
\renewcommand{\thepage}{A\arabic{page}}  
\renewcommand{\thesection}{A\arabic{section}}   
\renewcommand{\thetable}{A\arabic{table}}   
\renewcommand{\thefigure}{A\arabic{figure}}

\section*{Checklist}


\begin{enumerate}

\item For all authors...
\begin{enumerate}
  \item Do the main claims made in the abstract and introduction accurately reflect the paper's contributions and scope?
    \answerYes{}
  \item Did you describe the limitations of your work?
    \answerYes{Please see section~\ref{sec:data_model}.}
  \item Did you discuss any potential negative societal impacts of your work?
    \answerYes{Please see section~\ref{sec:conclusion}.}
  \item Have you read the ethics review guidelines and ensured that your paper conforms to them?
    \answerYes{}{}
\end{enumerate}

\item If you are including theoretical results...
\begin{enumerate}
  \item Did you state the full set of assumptions of all theoretical results?
    \answerNA{Our work does not contain theoretical results.}
	\item Did you include complete proofs of all theoretical results?
    \answerNA{Our work does not contain theoretical results.}
\end{enumerate}

\item If you ran experiments...
\begin{enumerate}
  \item Did you include the code, data, and instructions needed to reproduce the main experimental results (either in the supplemental material or as a URL)?
    \answerYes{We used publicly available data in all of our experiments. Meanwhile we either provide the detailed implementations or cite the papers of them following the authors instructions (See Section~\ref{sec:exp}). All of our codes are provided in \url{https://github.com/VITA-Group/SViTE}.}
  \item Did you specify all the training details (e.g., data splits, hyperparameters, how they were chosen)?
    \answerYes{We provides all the training details in Section~\ref{sec:exp}.}
	\item Did you report error bars (e.g., with respect to the random seed after running experiments multiple times)?
    \answerNo{ We did not report the error bars since running vision transformer on ImageNet are extremely resource-consuming. For example, each reported number takes around $960$ V100 GPU hours. We will continue running the experiments and report the confidence intervals in future versions.}
	\item Did you include the total amount of compute and the type of resources used (e.g., type of GPUs, internal cluster, or cloud provider)?
    \answerYes{We describe the details of computation resources in Section~\ref{sec:more_implementation} of the supplement.}
\end{enumerate}

\item If you are using existing assets (e.g., code, data, models) or curating/releasing new assets...
\begin{enumerate}
  \item If your work uses existing assets, did you cite the creators?
    \answerYes{We used publicly available data, i.e., ImageNet, in our experiments. We cited the corresponding papers published by the creators in Section~\ref{sec:exp}.}
  \item Did you mention the license of the assets?
    \answerNo{The license of ImageNet is included in the paper that we have cited.}
  \item Did you include any new assets either in the supplemental material or as a URL?
    \answerYes{The ImageNet we used are publicly available. And all our codes are included in \url{https://github.com/VITA-Group/SViTE}.}
  \item Did you discuss whether and how consent was obtained from people whose data you're using/curating?
    \answerNA{We did not collect/curate new data.}
  \item Did you discuss whether the data you are using/curating contains personally identifiable information or offensive content?
    \answerNA{All ImageNet datasets are already publicly available and broadly adopted. I do not think there are any issues of personally identifiable information or offensive content.}
\end{enumerate}

\item If you used crowdsourcing or conducted research with human subjects...
\begin{enumerate}
  \item Did you include the full text of instructions given to participants and screenshots, if applicable?
    \answerNA{}
  \item Did you describe any potential participant risks, with links to Institutional Review Board (IRB) approvals, if applicable?
    \answerNA{}
  \item Did you include the estimated hourly wage paid to participants and the total amount spent on participant compensation?
    \answerNA{}
\end{enumerate}

\end{enumerate}



\section{More Implementation Details} \label{sec:more_implementation}
\vspace{-2mm}
\paragraph{Computing resources.} All experiments use Tesla V100-SXM2-32GB GPUs as computing resources. Specifically, each experiment is ran with $8$ V100s for $4\sim5$ days.  

\paragraph{Why do we choose $600$ training epochs for SViTE experiments?} Choosing $600$ training epochs for sparse training is to maintain similar training FLOPs compared to dense ViT training. Specifically, if training a dense DeiT model for $300$ epochs needs $1$x FLOPs, training SViTE-DeiT at $40\%$ sparsity for $600$ epochs needs $\sim0.95$x FLOPs. In summary, we compare our SViTE ($600$ epochs) and DeiT baselines ($300$ epochs) based on a similar training budget. Such comparison fashion is widely adopted in sparse training literature like~\cite{liu2021we,evci2020rigging} (see Table $2$’s caption in~\cite{liu2021we} and Figure $2$ in~\cite{evci2020rigging} for details). Meanwhile, note that the reported running time is per epoch saving (i.e., total running time / total epoch), which would not be affected by the number of training epochs.

\paragraph{Baseline models with longer epochs.} Actually, the performance of DeiT training without distillation saturates after $300\sim400$ epochs, as stated in~\cite{touvron2020training}. We also conduct longer epoch ($600$ epochs) training for DeiT-Small and -Base models. Our results collected in the table~\ref{tab:longer_epochs} align with the original DeiT paper~\cite{touvron2020training}. It suggests that our proposed SViTE is still able to achieve better accuracy with fewer parameters and fewer training\&inference computations. Specifically, at $40\%$ structured sparsity, our sparsified DeiT-Base can achieve $0.21\%$ accuracy gain, at $\sim51\%$ training FLOPs, $33.13\%$ inference FLOPs, and $24.70\%$ running time savings, compared to its dense counterpart with $600$ epochs.

\begin{table}[!ht]
\caption{\small Performance of DeiT-Small/-Base with longer training epochs on ImageNet-1K.}
\label{tab:longer_epochs}
\centering
\resizebox{1\textwidth}{!}{
\begin{tabular}{l|cccc}
\toprule
\multirow{1}{*}{Metrics} & \multirow{1}{*}{DeiT-Small $300$ Epochs} & \multirow{1}{*}{DeiT-Small $600$ Epochs} & \multirow{1}{*}{DeiT-Base $300$ Epochs} & \multirow{1}{*}{DeiT-Base $600$ Epochs}  \\  
\midrule
Accuracy (\%) & 79.90 & 80.02 \scalebox{0.8}{(+0.12)} & 81.80 & 82.01 \scalebox{0.8}{(+0.21)}\\ 
\bottomrule
\end{tabular}}
\vspace{-4mm}
\end{table}

\section{More Experimental Results}
\begin{figure}[!ht]
    \centering
    \includegraphics[width=1\linewidth]{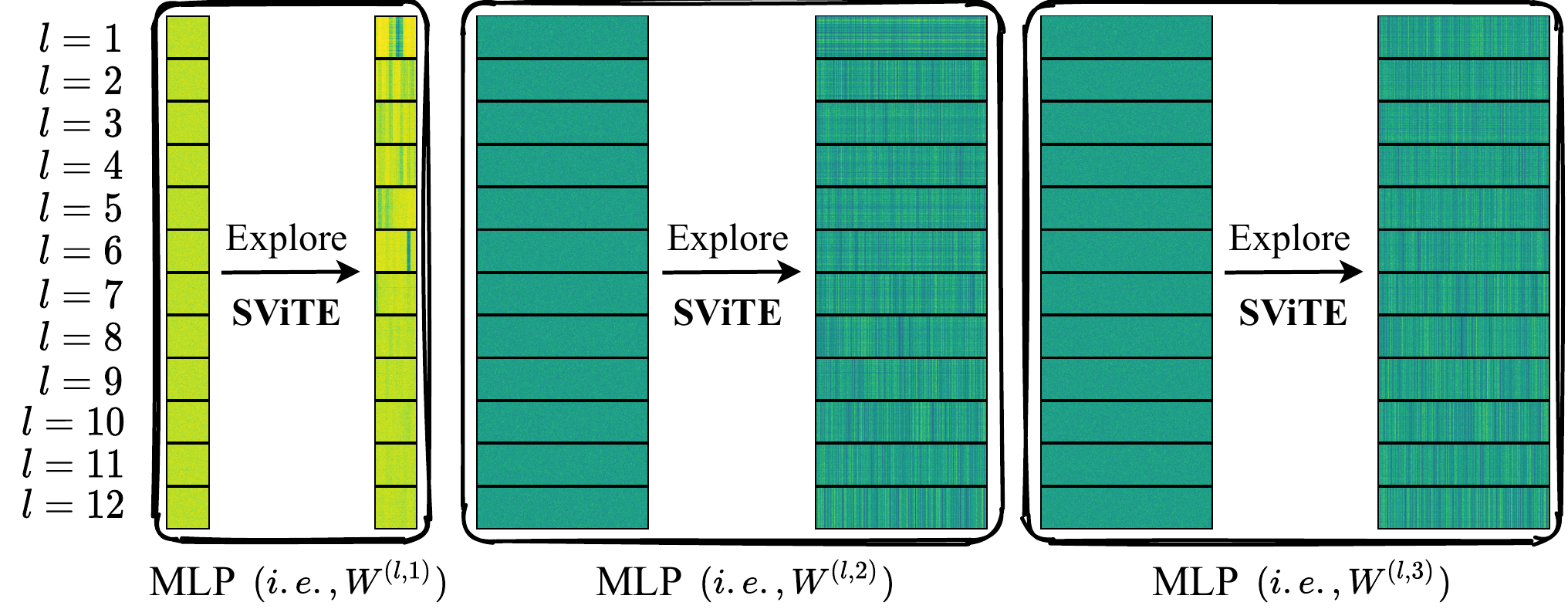}
    \vspace{-6mm}
    \caption{\small Binary mask visualizations of SViTE-Base at $50\%$ unstructured sparsity. Within each box, \textit{left} is the initial random mask; \textit{right} is the explored mask from SViTE.}
    \label{fig:mask_uns}
    \vspace{-4mm}
\end{figure}

\vspace{-2mm}
\paragraph{Sparse topology of SViTE-Base with unstructured sparsity.} As shown in Figure~\ref{fig:mask_uns}, we observe that from the initial random mask to explored mask in SViTE, plenty of structural patterns emerge (i.e., the darker ``vertical" lines mean completely pruned neurons in the MLPs). It is supervising that unstructured sparse exploration can lead to structured patterns, which implies the great potential to be accelerated in real-world hardware devices.

\begin{figure}[!ht]
    \centering
    \includegraphics[width=1\linewidth]{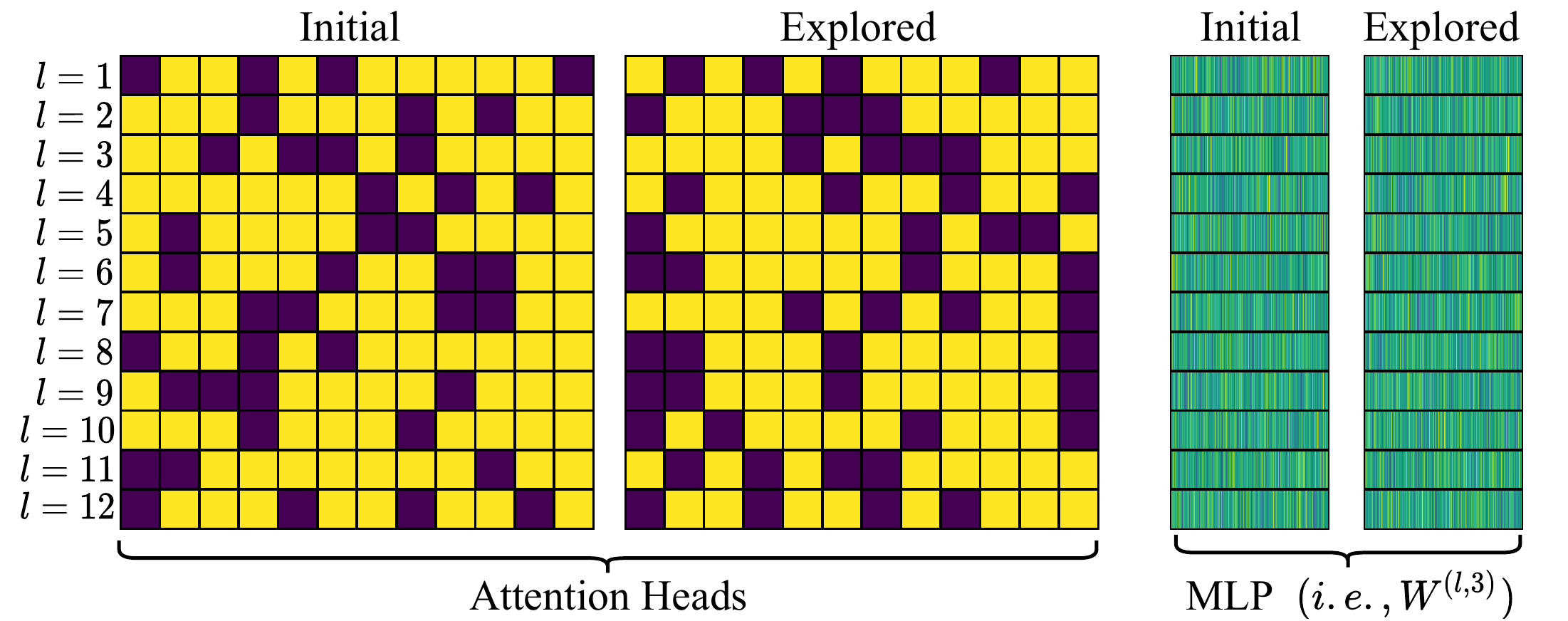}
    \vspace{-6mm}
    \caption{\small (\textit{Left}) The ``survivors" summary of existing attention heads in sparse vision transformers from S$^2$ViTE. Dark entry is the pruned attention head; bright entry means the remaining attention head. (\textit{Right}) Binary masks of all $W^{(l,3)}$ MLPs in S$^2$ViTE-Base. \textit{Initial} denotes the random connectivity in the beginning, and \textit{Explored} is the explored typologies at the end. Visualized S$^2$ViTE-Base has $40\%$ structural sparsity.}
    \label{fig:mask_s}
    \vspace{-4mm}
\end{figure}

\paragraph{Sparse topology of S$^2$ViTE-Base with structured sparsity.} Figure~\ref{fig:mask_s} shows mask visualizations of pruned multi-attention heads and MLPs in vision transformers. It shows that S$^2$ViTE indeed explores totally different connectivity patterns, compared to the initial topology.

\paragraph{Ablation of only applying our learnable token selector.} We compare these three setup: (a) DeiT-Small ($79.90$ test accuracy); (b) DeiT-Small + Token selector with $10\%$ data sparsity ($78.67$ test accuracy); (c) DeiT-Small + Token selector with $10\%$ data sparsity + SViTE with $50\%$ unstructured sparsity ($79.91$ test accuracy). It demonstrates that simultaneously enforcing data and architecture sparsity brings more performance gains. 

\paragraph{Ablation of the layerwise sparsity of attention maps.} As shown in Table~\ref{tab:sparse_att}, we investigate the layerwise sparsity of attention maps. Dense DeiT-Small and SViTE+-Small with $10\%$ data sparsity and $50\%$ model sparsity are adopted for experiments. We calculate the percentage of elements in attention maps whose magnitude is smaller than $10^{-4}$. We observe that the bottom layers' attention maps of SViTE+ are denser than the ones in dense ViT, while it is opposite for the top layers. 

\begin{table}[!ht]
\caption{\small Layerwise sparsity of attention maps.}
\label{tab:sparse_att}
\centering
\resizebox{1\textwidth}{!}{
\begin{tabular}{l|cccccccccccc}
\toprule
\multirow{1}{*}{Layer Index} & \multirow{1}{*}{1} & \multirow{1}{*}{2} & \multirow{1}{*}{3} & \multirow{1}{*}{4} & 5 & 6 & 7 & 8 & 9 & 10 & 11 & 12 \\  
\midrule
Dense-Small  & 28.44 & 44.70 & 37.04 & 13.87 & 13.62 & 13.67 & 12.98 & 5.99 & 5.08 & 3.46 & 2.74 & 15.05\\ 
SViTE+-Small & 24.37 & 33.24 & 28.05 & 19.51 & 9.03 & 13.79 & 13.24 & 8.88 & 6.78 & 5.65 & 2.88 & 35.22 \\ 
\bottomrule
\end{tabular}}
\vspace{-4mm}
\end{table}

\end{document}